\documentclass{Interspeech}

\interspeechcameraready

\title{Data Augmentation for Spoken Grammatical Error Correction}

\author[affiliation={}]{Penny}{Karanasou}
\author[affiliation={}]{Mengjie}{Qian}
\author[affiliation={}]{Stefano}{Bannò}
\author[affiliation={}]{Mark J.F.}{Gales}
\author[affiliation={}]{Kate M.}{Knill}

\affiliation{ALTA Institute, Department of Engineering}{University of Cambridge}{UK}
\email{\{pk407,mq227,sb2549,mjfg100,kmk1001\}@cam.ac.uk}


\usepackage{comment}
\usepackage{amsmath, multirow, multicol}
\usepackage{subcaption}

\usepackage{comment,makecell}
\usepackage{tabularx, url}
\usepackage[table]{xcolor}
\definecolor{gray}{rgb}{0.5,0.5,0.5}
\definecolor{Gray}{gray}{0.9}

\begin{document}

\maketitle

\footnotetext[1]{This paper reports on research supported by Cambridge University Press \& Assessment, a department of The Chancellor, Masters, and Scholars of the University of Cambridge.}

\begin{abstract}    
    While there exist strong benchmark datasets for grammatical error correction (GEC), high-quality annotated spoken datasets for Spoken GEC (SGEC) are still under-resourced. In this paper, we propose a fully automated method to generate audio-text pairs with grammatical errors and disfluencies. Moreover, we propose a series of objective metrics that can be used to evaluate the generated data and choose the more suitable dataset for SGEC. The goal is to generate an augmented dataset that maintains the textual and acoustic characteristics of the original data while providing new types of errors. This augmented dataset should augment and enrich the original corpus without altering the language assessment scores of the second language (L2) learners. We evaluate the use of the augmented corpus both for written GEC (the text part) and for SGEC (the audio-text pairs). Our experiments are conducted on the S\&I Corpus, the first publicly available speech dataset with grammar error annotations.
\end{abstract}

\section{Introduction}

Automatic spoken language assessment (SLA) is the task of grading second language (L2) learners and providing them feedback in an automatic way, without human expert knowledge. Grammatical error correction (GEC) is an important part of SLA and a well-established research area \cite{bryant2023}, supported by a number of shared tasks such as
CoNLL-2014 \cite{ng2014conll}, BEA-2019~\cite{bryant2019bea}, and MULTIGEC-2025 \cite{multigec2024}. These efforts have led to strong benchmark datasets and models capable of correcting a wide range of grammatical errors in text. However, this is not the case for Spoken GEC (SGEC), where data are still sparse and there are different challenges to address. Spoken data are inherently noisy and contain disfluencies such as hesitations, repetitions, and false starts, as well as incomplete or fragmented sentences, accented speech, and the lack of punctuation and capitalization. These factors make it significantly more difficult to detect and correct grammatical errors compared to written text. They also complicate the annotation process, making it more time-consuming and costly. 

\begin{figure}[!t]
    \centering
    \includegraphics[width=0.95\linewidth]{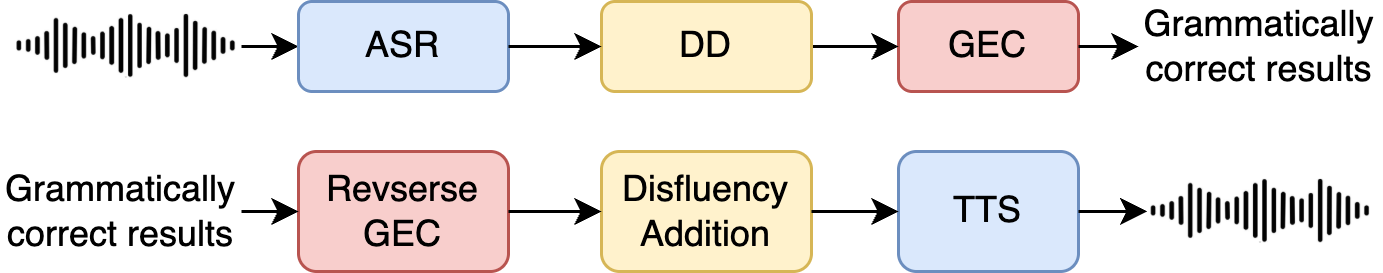}
    \caption{Cascaded spoken GEC process (top) and automatic spoken GEC data generation process (bottom).}
    \label{fig:sgec_reverse-gec}
\end{figure}

A solution to this is to apply data augmentation technique to avoid the manual effort of creating and annotating a new corpus. Traditionally, SGEC systems have followed a cascaded pipeline \cite{lu2020spoken}, starting with an automatic speech recognition (ASR) module to transcribe audio into text, followed by a disfluency detection (DD) module to generate fluent transcriptions, and finally a GEC module to correct the grammar. 
\cite{qian2025scaling,qian2025sgec-journal} leverage largely available audio recordings to generate pseudo GEC transcriptions using a semi-cascaded pipeline with Whisper and a text-based GEC model.
In this paper, a reverse pipeline is proposed to automatically generate data by introducing grammatical errors and disfluencies into clean text. Our pipeline includes a reverse GEC module, a module for disfluencies addition, and finally a Text-to-Speech (TTS) module to generate aligned audio from the augmented text (Figure \ref{fig:sgec_reverse-gec}). 

Previous research has explored different parts of this process. For reverse GEC, early approaches include the use of edit distance, word embeddings and spell-breaking~\cite{grundkiewicz-junczys-dowmunt-2019-minimally}, as well as simple perturbations like misspelling, word swapping and reversing~\cite{10.1016/j.jksuci.2023.101572}. Another family of approaches uses deep learning to generate pseudo-data for Grammatical Error Correction (GEC). For example, \cite{koyama2021comparison} compares different back-translation models, including Transformers, CNNs, and LSTMS. Lately, Large Language Models (LLMs) have also been used for GEC and reverse GEC. \cite{luhtaru2024toerr} uses LLaMA models for GEC and reverse GEC for German, Estonian and Ukrainian. 
Similar results are achieved with prompting GPT3.5 vs fine-tuning Llama. 

Concerning the addition of disfluencies, some rule-based approaches are proposed, such as in \cite{passali2025artificial}. However, in this case the types of disfluencies are limited to the type of included rules and their frequency does not necessarily match the frequency of appearance in real data. Using BERT-based models to add disfluencies is another proposed approach~\cite{dubov2023thesis}. 

Once the transcription has been augmented with errors and disfluencies, a TTS system can be used to generate audio. TTS has been successfully used for data augmentation in ASR~\cite{huang2023textgenerationspeechsynthesis,Fazel2021}. In our case, the TTS system must meet several constraints: it should support voice cloning, preserve speaker characteristics (i.e. accented speech), generate well-aligned speech with the target text, and deliver high audio quality. In this paper, different TTS systems are investigated in this direction.

In our paper, the contribution is threefold. First, we present an automated pipeline for data augmentation that includes a model-based reverse GEC module, a disfluencies addition module and a TTS module. Second, we present four objective metrics that can evaluate the quality of the generated data. These metrics offer a comprehensive analysis of the acoustic and textual characteristics of the generated data, and they allow us to choose the best suited dataset for SGEC from a set of available models. Finally, we evaluate our augmented dataset in the GEC and SGEC tasks. We are working with the Speak \& Improve Corpus 2025 \cite{knill2024speakimprovecorpus}, the first publicly available speech dataset with grammar error annotations. 

\section{Data Augmentation Pipeline}

We present a model-based data augmentation pipeline that generates aligned text and audio data with added grammatical errors and disfluencies. 
Each module is detailed below.

\subsection{Reverse GEC Module}
\label{sec:reverseGEC}

The first step adopts a reverse GEC module to automatically generate text with grammatical errors. A BART-large model~\cite{lewis2020bart}, previously used for GEC in the S\&I 2025 Challenge~\cite{qian2024speakimprovechallenge}, is used with the same hyperparameters, except with the input and output reversed. Specifically, the input to the model is manual GEC transcriptions (i.e. text without grammatical errors) and the output is corresponding original transcriptions containing errors. There are $\sim1M$ utterances from the BEA-2019 set used for the GEC model as in~\cite{qian2024speakimprovechallenge}. However, only half of these have grammatical errors. 
We observe that training solely on utterances with errors reinforces the model's ability to generate diverse and meaningful errors. Although this reduces the training size to $\sim 500k$, it more than doubles the number of outputs containing new errors, i.e. errors not present in the original transcriptions. A checkpoint is selected that produces new errors in over $80\%$  of the generated utterances.

\begin{figure}[h]
\centering
\includegraphics[trim=0 5mm 0 0, width=8cm, height=5cm]{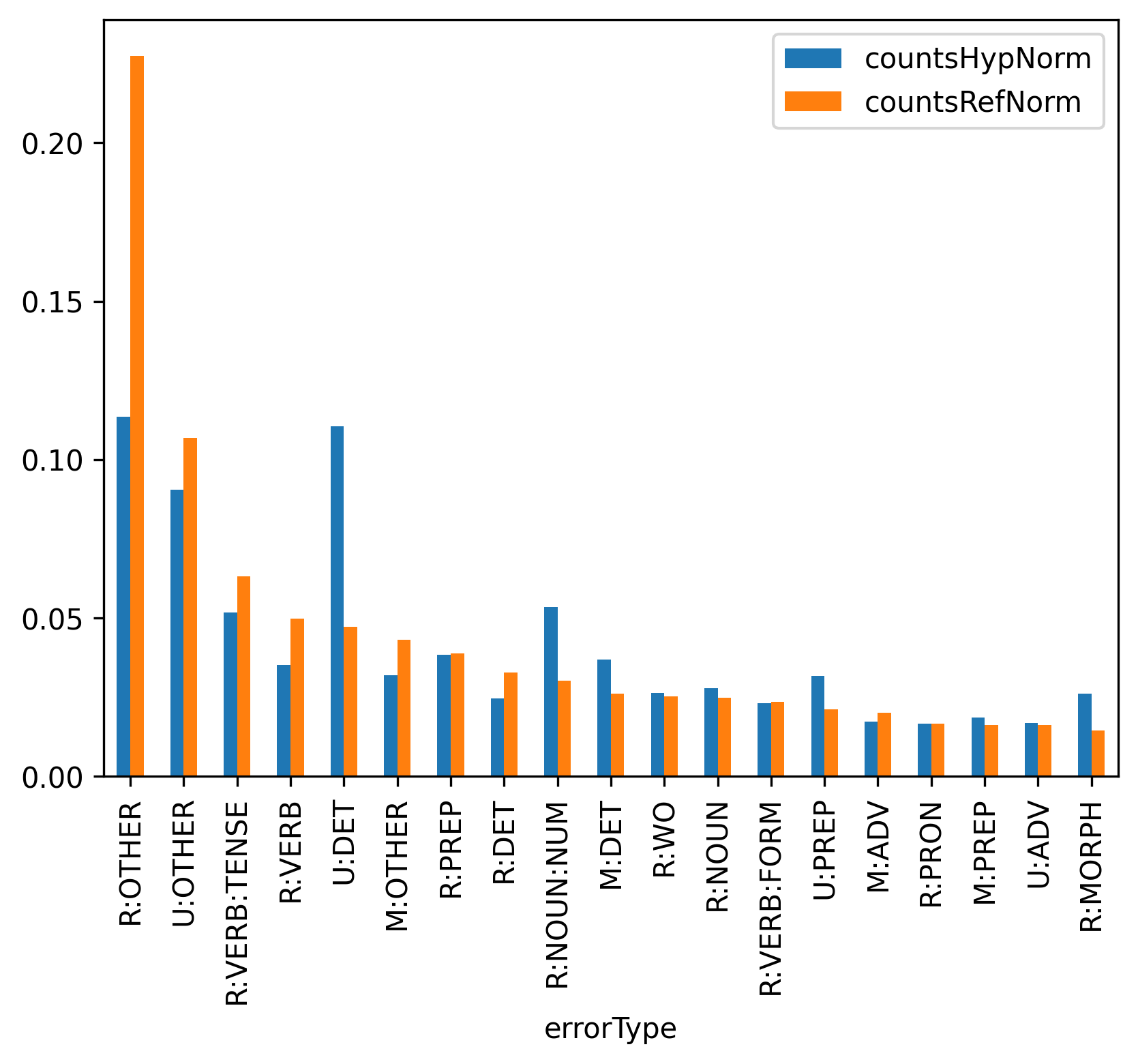}
\caption{Distribution plot of error categories in ref (original transcriptions) and hyp (transcriptions with generated errors)}
\label{fig:distribution_plot}
\end{figure}

To ensure the quality and relevance of generated errors, several sanity checks are conducted on a randomly sampled set of training utterances (6k). Figure~\ref{fig:distribution_plot} presents the 20 most frequent error types, comparing their normalized occurrence in the reference (original transcriptions) and hypothesis (generated error) sets. Overall, the distributions follow similar trends. Notable differences include higher frequencies of generated errors in determinants and numbers, which are categories relatively easy for the model to learn. In contrast, the ``Other'' category shows fewer errors, as it includes random errors that are harder to generalize and more difficult for a reverse GEC model to learn.

As an extra sanity check, the generated transcriptions from the reverse GEC are passed through the GEC system trained for the S\&I 2025 Challenge. This is to verify that the generated errors can be corrected by the existing GEC system, indicating that they align with the error types seen in its training set. Table~\ref{tab:table-errant-GEC} reports the F$_\text{0.5}$ score calculated based on the ERRANT edits~\cite{bryant-etal-2017-automatic}. Interestingly, the score on the generated transcriptions outperforms the original transcriptions, giving a strong indication that the generated errors are of the expected types.


\begin{table}[!htbp]
  \centering
  \caption{ERRANT-based Precision, Recall, and F$_\text{0.5}$ scores for original vs. reverse GEC-generated transcriptions.}
 \label{tab:table-errant-GEC}
    \begin{tabular}{l|ccc} 
     \toprule
      \textbf{Text} & \textbf{P} & \textbf{R} & \textbf{F$_\text{0.5}$}\\
      \midrule
      Original & 65.4 & 24.9 & 49.3\\
      Reverse GEC & 77.6 & 26.1 & 55.6\\
      \bottomrule
    \end{tabular}
\end{table}

\subsection{Disfluencies Addition Module}

From this point onward, the S\&I corpus (see Section \ref{sec:corpus}) is used, which includes text and audio annotated with grammatical errors and disfluency tags. The disfluencies included are hesitations, repetitions, false starts and incomplete sentences, which offer a good coverage of typical spoken disfluencies. Each file is aligned with the output of the reverse GEC system, allowing disfluencies to be inserted at their original positions. This approach preserves the frequency of disfluency occurrences while minimizing changes to the text. We hypotheses that minimizing text edits may ease voice cloning in the TTS step by keeping the augmented transcription closer to the original. Exploring how TTS systems handle varying levels of inserted errors and disfluencies, as well as developing model-based or rule-based disfluency insertion methods, is left for future work.

\subsection{TTS module}

Once augmented transcriptions with disfluencies are ready, they are used as target text for speech synthesis. The goal is to preserve the original speaker's voice while modifying the transcription. The TTS must be able to accurately clone the speaker's accent and make the necessary edits to the target text. This process will yield aligned speech-text pairs suitable for end-to-end SGEC model training.
Our chosen TTS is a pretrained F5-TTS model~\cite{chen2024f5ttsfairytalerfakesfluent}, a flow-matching non-autoregressive TTS system based on Diffusion Transformer~\cite{peebles2023scalable-diffusion}. Trained on a $100K$ hours multilingual dataset, F5-TTS demonstrates natural and expressive zero-shot capabilities and offers faster inference compared to other diffusion-based models. 
Before choosing F5-TTS, several other TTS systems are informally evaluated through subjective listening. All systems that we experimented with are multi-lingual and multi-speaker, with zero-shot voice cloning capabilities. Pretrained models are used without fine-tuning. VoiceCraft~\cite{peng2024voicecraft}, which supports both TTS and speech editing, initially showed promise. However, its inference speed was too slow for even medium-scale datasets. In editing mode, the system requires specifying the type of edit (substitution, deletion, insertion), limiting automation, especially for longer or more complex inputs. It also exhibited inconsistencies in handling longer sentences or certain edit types. VALL-E-X~\cite{zhang2023speakforeignlanguagesvoice} seemed to successfully produce the correct transcription, but it failed to transfer the speaker's identity, resulting in a generic and somewhat robotic voice. Both YourTTS~\cite{casanova2022yourtts} and XTTS-v2~\cite{casanova24xtts} performed reasonably well in generating accurate transcriptions and transferring the source voice, but produced lower audio quality compared to F5-TTS. Orpheus-TTS~\cite{canopy2025orpheus-tts} failed to generate either the transcription or the voice effectively for our use case.


\section{Objective Metrics for Spoken Augmented Data Analysis}
\label{sec:data_metrics}

This section proposes four metrics to evaluate and compare different versions of the generated spoken augmented data. These metrics offer a systematic approach to selecting the optimal set of spoken augmented data from multiple candidates. In this study, we will only change the TTS model used in the data augmentation framework and compare two main alternatives, F5-TTS and VoiceCraft (in both TTS and Speech Editing modes). However, the same metrics could be applied if any part of the pipeline changed, such as the reverse GEC or disfluency addition modules. Future work will explore modifications to these parts of the pipeline. 
The goal is to assess whether the augmented data preserve the acoustic characteristics of the source speaker and the semantics of the original transcription, ensuring that the augmented data yield comparable assessment scores to those obtained from the original ones. All metrics are computed on a subset of the S\&I Corpus dev set ($\sim 3300K$ utterances). 


\subsection{Speaker Verification}
\label{sec:spkID}

\begin{figure}[!htbp]
\centering
\includegraphics[trim=0 8mm 0 0mm, width=0.85\linewidth]{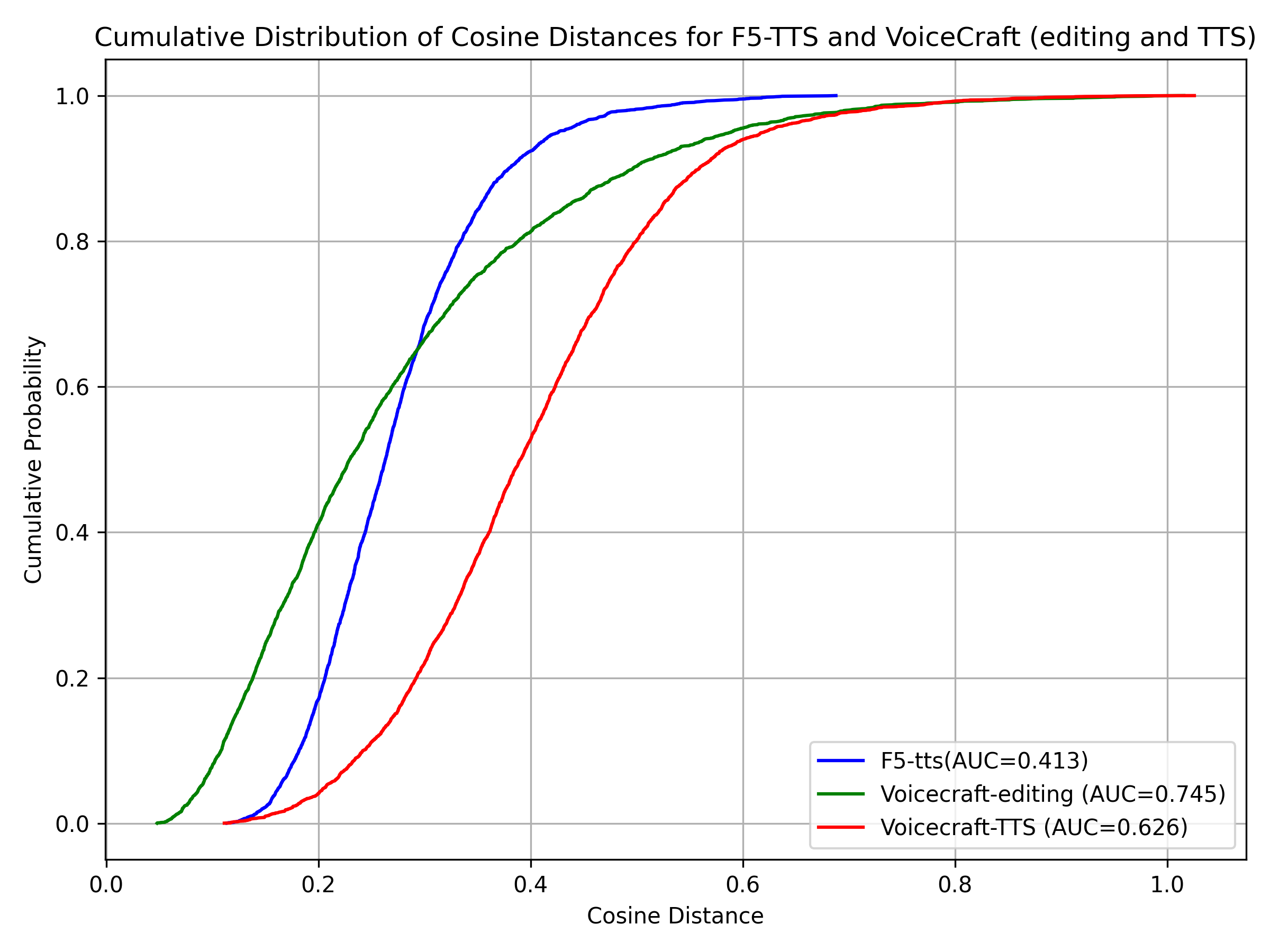}
\caption{Cumulative plots and AUC for cosine distances for F5-TTS and VoiceCraft (Editing and TTS mode)}
\label{fig:cumulative}
\end{figure}

To ensure the speaker identity is preserved in the augmented data, speaker embeddings are extracted using Pyannote~\cite{bredin2023pyannote}. Cosine distance between the speaker embeddings of the original and generated speech is then computed as a similarity metric, where $0.0$ indicates perfect similarity and $2.0$ the worst case.
Figure~\ref{fig:cumulative} presents the cumulative plots of cosine distances for the systems under comparison. The F5-TTS curve is the steepest on the left side, indicating a higher proportion of closely matched pairs between original and generated data. This observation is further supported by the Area Under the Curve (AUC), where F5-TTS yields the lowest value. A lower AUC value indicates that the distance values are clustered towards the lower end of the range, thereby demonstrating greater similarity in speaker identity between the original and generated speech.

\subsection{ASR Output}


We calculate the Word Error Rate (WER) of an ASR system on pairs of original audio and original transcriptions, as well as on pairs of generated transcriptions and audio generated from F5-TTS and VoiceCraft in Speech Editing and TTS mode. This metric assesses how closely the generated audio matches its corresponding generated transcription. A low WER indicates better audio-text alignment, and thus more suitable data for GEC and SGEC training. 
For ASR evaluation, we use a Whisper$_\text{dsf}$ small.en model trained on Linguaskill data~\cite{banno2024towards,qian2025scaling}, this model produces disfluent transcriptions. Note that WER also reflects ASR system errors. We assume these are consistent with the baseline WER (from original audio/transcription pairs), so any increase is attributed to errors introduced by the TTS/Speech Editing systems.
WER results are shown in Table~\ref{tab:wer}. The WER of F5-TTS is the closest to the Original baseline, while VoiceCraft, both its TTS and Speech Editing mode, presents a significant increase in errors. We notice an increase in insertions in the F5-TTS WER. Upon manual error inspection, this appears to be primarily due to a few prompt words being included at the start of the generated audio.


\begin{table}[!t]
    \centering
    \caption{WERs on different types of audio samples.}
    \label{tab:wer}
    \begin{tabular}{l|cccc}
    \toprule
     \textbf{Dataset}  &  \textbf{WER} &  \textbf{Sub} &  \textbf{Del} &  \textbf{Ins} \\
    \midrule
    Orig   & 8.3 & 5.1 & 1.9 & 1.4      \\
      \cmidrule{1-5}
    VoiceCraft-Edited & 21.1 & 10.4 & 5.9 & 4.8\\
    VoiceCraft-TTS &  17.6 & 9.2 & 4.1 & 4.2 \\
    F5-TTS & 12.0 & 2.2 & 1.1 & 8.7 \\
    \bottomrule
    \end{tabular}
\end{table}

\subsection{BERT text-based and Wav2vec audio-based SLA Graders}

\begin{figure}[!htbp]
    \centering
    \begin{subfigure}[b]{\linewidth}
        \centering
        \includegraphics[width=6.5cm, height=4.5cm]{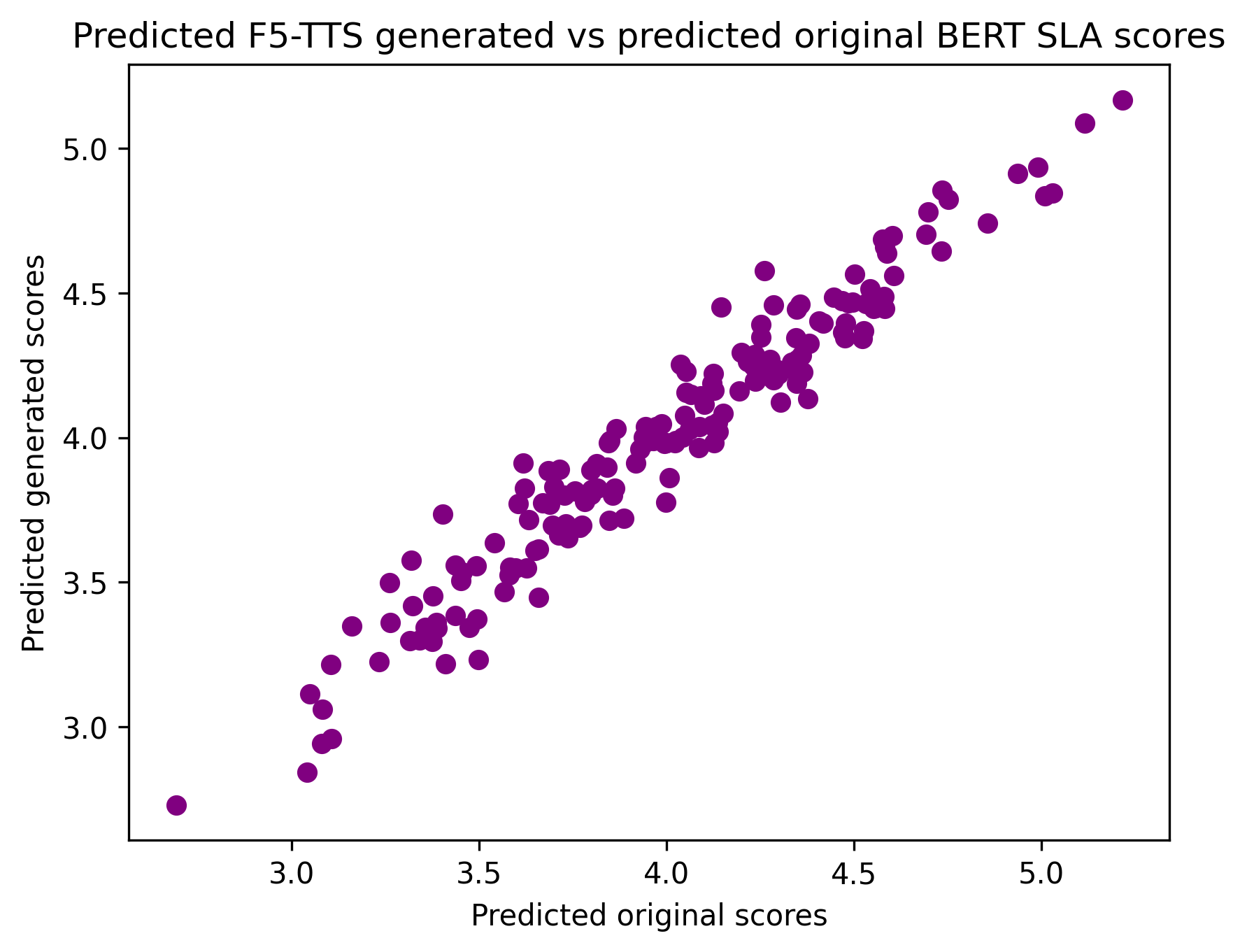}
    \end{subfigure}



    \hfill
    \begin{subfigure}[b]{\linewidth}
        \centering
       \includegraphics[width=6.5cm, height=4.5cm]{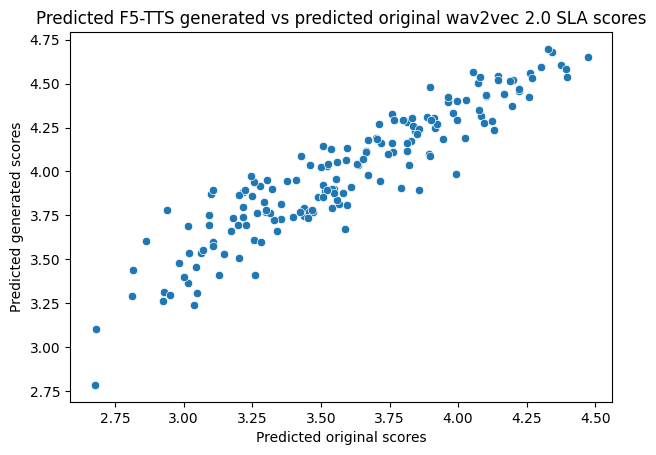}
    \end{subfigure}
   
    \caption{Scatter plots of BERT and wav2vec 2.0 SLA scores: Predicted scores from generated data vs. original data.}
    \label{fig:scatterplots}
\end{figure}

\noindent
The next two proposed metrics use text-based and audio-based Spoken Language Assessment (SLA) graders. Ideally, the generated data should yield SLA scores similar to the original ones, indicating that the augmented corpus does not alter learner assessment outcomes.
For text scoring, we use a BERT-based SLA grader trained on Linguaskill data~\cite{ludlow2020official}. The input to the grader includes the original reference text (Orig), ASR output of Whisper$_\text{dsf}$ for the original audio (Orig-ASR), the augmented data predicted from reverse GEC with added disfluencies (Generated-text) and the output of Whisper$_\text{dsf}$ for the generated TTS audio (VoiceCraft(VC)-Editing-ASR, VC-TTS-ASR, F5-TTS-ASR). The final scores are computed using an ensemble of 10 seeds with calibration per part and averaging the per part results. The BERT embeddings are expected to capture both syntactic and semantic textual features.
For audio scoring, we use a wav2vec2-based grading system trained on Linguaskill data~\cite{ludlow2020official} on Original audio and on TTS generated audio. The grader operates on raw waveform and produces latent representations that capture acoustic features such as pitch, duration, formants, and timbre. It may also capture speaker-specific characteristics of speech, semantic information and temporal patterns.  Smaller differences in embeddings suggest greater preservation of the original acoustic characteristics in the generated speech. 

Figure~\ref{fig:scatterplots} presents scatter plots comparing F5-TTS predicted scores with original scores, showing strong positive correlations for both BERT (top) and wav2vec2 (bottom) scores. This indicates that the generated augmented data is of the expected quality and doesn't alter the assessment scores. It also suggests that the SLA graders demonstrate some robustness to newly introduced errors. Similar patterns were observed when comparing VoiceCraft-TTS and VoiceCraft-Edited predicted scores against Original predicted scores, as well as Original ASR predicted scores against Original manual transcriptions, and Generated text against Original manual transcriptions. These additional plots have been omitted here due to space constraints.

\section{Spoken GEC evaluation}

\subsection{Corpus}
\label{sec:corpus}

The Speak \& Improve (S\&I) Corpus 2025 is a comprehensive, public dataset of second-language (L2) learner speech, designed to facilitate research in automated spoken language assessment and feedback~\cite{knill2024speakimprovecorpus}. Collected through the Speak \& Improve platform between 2019 and 2024~\cite{nicholls23_interspeech}, the corpus features diverse learner audio recordings with detailed annotations including manual transcriptions, disfluency markings, grammatical corrections, and Common European Framework of Reference (CEFR)~\cite{cefr2001} proficiency ratings ranging from A2 to C. The S\&I Corpus comprises five distinct parts, each targeting specific aspects of speaking proficiency: Part 1 involves answering short questions with timed responses; Part 2 is a read-aloud task featuring sentence reading; Parts 3 and 4 require 1 min talk to provide candidates' opinion on a given topic or describe the process depicted in the given diagram.; and Part 5 includes responses to 5 questions related to a topic, each question is around 20 seconds long. The scores for each part range from 2.0 to 6.0, approximately in correlation with CEFR levels A2 to C2, and an average of these scores determines the overall speaking proficiency of the candidate. This corpus has been released as part of the S\&I Challenge 2025 \cite{qian2024speakimprovechallenge} (parts 1,3-5 only). For SGEC model training, 5 hours of the dev set are used for validation, and the rest is merged with the training set. 

\subsection{Evaluation of Written and Spoken Augmented Data}
\label{sec:gec_eval}

\begin{table}[!htbp]
    \centering
    \caption{WER and ERRANT scores of the S\&I eval set with written augmented data for different cascaded SGEC models.}
    \label{tab:writtenGEC}
    \begin{tabular}{@{ }p{6.4mm}@{ }|p{6.2mm}|@{ }p{5.4mm}p{5.4mm}p{5.6mm}}
    \toprule
    Model & WER & P & R & F$_\text{0.5}$ \\
    \cmidrule{1-5}
    \multicolumn{1}{l|}{Manual + GEC}  & \multirow{2}*{0.00} & 68.03  & 26.79 &  52.01\\
    \multicolumn{1}{l|}{Manual + GEC-Aug} & & 63.42 & 35.79 & \textbf{54.94}\\
    \midrule
    \cmidrule{1-5}
    \multicolumn{1}{l|}{Whs$_\text{small}$ + DD + GEC }  & 17.73  & 37.18 & 8.44 & 22.11\\
    \multicolumn{1}{l|}{Whs$_\text{small}$ + DD + GEC-Aug}  & 17.57 & 35.74 & 15.42 & \textbf{28.28}\\
    \hline
    \multicolumn{1}{l|}{Whs$_\text{large}$ + DD + GEC}  & 15.66 & 45.95 & 8.47 & 24.37  \\
    \multicolumn{1}{l|}{Whs$_\text{large}$ + DD + GEC-Aug} & 15.72 & 39.40 & 15.76 & \textbf{30.31} \\
    \hline
     \multicolumn{1}{l|}{Whs$_\text{flt}$  + GEC}  & 14.82 & 51.78 & 16.11 & 35.89 \\
     \multicolumn{1}{l|}{Whs$_\text{flt}$  + GEC-Aug}  & 14.68 & 45.42 & 24.35 & \textbf{38.72}\\
      \bottomrule
    \end{tabular}
\end{table}

\begin{table}[!htbp]
    \centering
    \caption{WER and ERRANT scores of the S\&I eval set with spoken augmented data for end-to-end SGEC models (Whs$_\text{gec}$).}
    \label{tab:SpokenGEC}
    \begin{tabular}{@{ }l@{ }|p{6.2mm}|p{4.6mm}p{4.6mm}p{4.8mm}}
    \toprule
   Training \& Fine-tuning Data & WER & P & R & F$_\text{0.5}$ \\
    \midrule
    Train+Dev & 12.75 & 43.89 & 33.70 & 41.39 \\
    Train+Dev+Gen  &  12.75 & 43.70 & 33.98 & 41.34\\
    Train+Dev+Gen (cos04) & 12.69 & 44.04 & 34.21 & 41.65\\
    \quad + Fine-tuned on Train+Dev & 12.57 & 44.03 & 35.22 & \textbf{41.93} \\
    \bottomrule
    \end{tabular}
\end{table}

The reverse GEC system (Section~\ref{sec:reverseGEC}) is applied to the full GEC training set ($\sim 1M$ utterances), generating $\sim 870K$ utterances with new grammatical errors. 
These are combined with the original training data to form a text-based augmented training set, used to train a GEC model (GEC-Aug) with the same development set and hyperparameters as the original. Evaluation is performed on the S\&I eval set and compared to the S\&I Challenge baseline~\cite{qian2024speakimprovechallenge} (third row in Table \ref{tab:writtenGEC}). The first two rows of Table~\ref{tab:writtenGEC} show that GEC-Aug outperforms the baseline when applied to manual fluent transcriptions. GEC-Aug is then integrated into both cascaded and semi-cascaded SGEC pipelines (row 3 - 8). We compare three systems: Whs$_\text{small}$ and Whs$_\text{large}$, which are pretrained Whisper models used in a cascaded SGEC pipeline with DD and GEC; and Whs$_\text{flt}$, a model fine-tuned on fluent S\&I data with grammatical errors retained and disfluencies removed, used in a semi-cascaded setup. Training details follow the same setup as in See~\cite{qian2025scaling}.
Results indicate that augmented written data improves performance across 
all three models.

Next, audio-text augmented data are generated with the F5-TTS model and used to train an end-to-end SGEC system (Whs$_\text{gec}$)~\cite{banno2024towards}. We train a Whisper model with the original S\&I Corpus data (row Train+Dev in Table~\ref{tab:SpokenGEC}), and another model with additional generated audio-text data (row Train+Dev+Gen). To improve data quality, filtering is applied based on speaker similarity using the cosine distance from the speaker verification analysis of Section \ref{sec:spkID} (row Train+Dev+Gen (cos04)). This model (row 3) is further fine-tuned on the original S\&I Corpus audio-text data (row 4). The results show that using augmented data improves performance, with speaker-based filtering yielding further gains. Fine-tuning on the original data enhances performance even more.

\section{Conclusion}
In this work, we propose an automatic pipeline to generate audio-text pairs for training end-to-end SGEC models. This pipeline consists of a reverse GEC model, a disfluency addition module and a TTS component. Four metrics are then proposed to evaluate the quality of spoken augmented data and to select the optimal set of generated data. 
Various TTS models are compared with subjective listening and with the proposed metrics. 
The resulting augmented data improves performance when used to train both text-based GEC and SGEC tasks across cascaded, semi-cascaded, and end-to-end pipelines.

\newpage
\bibliographystyle{IEEEtran}
\bibliography{mybib}

\begin{thebibliography}{10}
\providecommand{\url}[1]{#1}
\csname url@samestyle\endcsname
\providecommand{\newblock}{\relax}
\providecommand{\bibinfo}[2]{#2}
\providecommand{\BIBentrySTDinterwordspacing}{\spaceskip=0pt\relax}
\providecommand{\BIBentryALTinterwordstretchfactor}{4}
\providecommand{\BIBentryALTinterwordspacing}{\spaceskip=\fontdimen2\font plus
\BIBentryALTinterwordstretchfactor\fontdimen3\font minus \fontdimen4\font\relax}
\providecommand{\BIBforeignlanguage}[2]{{%
\expandafter\ifx\csname l@#1\endcsname\relax
\typeout{** WARNING: IEEEtran.bst: No hyphenation pattern has been}%
\typeout{** loaded for the language `#1'. Using the pattern for}%
\typeout{** the default language instead.}%
\else
\language=\csname l@#1\endcsname
\fi
#2}}
\providecommand{\BIBdecl}{\relax}
\BIBdecl

\bibitem{bryant2023}
C.~Bryant, Z.~Yuan, M.~R. Qorib, H.~Cao, H.~T. Ng, and T.~Briscoe, ``{Grammatical Error Correction: A Survey of the State of the Art},'' \emph{Computational Linguistics}, vol.~49, no.~3, pp. 643--701, 09 2023.

\bibitem{ng2014conll}
\BIBentryALTinterwordspacing
H.~T. Ng, S.~M. Wu, T.~Briscoe, C.~Hadiwinoto, R.~H. Susanto, and C.~Bryant, ``{The CoNLL-2014 Shared Task on Grammatical Error Correction},'' in \emph{Proceedings of the 18th conference on computational natural language learning: shared task}, 2014, pp. 1--14. [Online]. Available: \url{https://aclanthology.org/W14-1701/}
\BIBentrySTDinterwordspacing

\bibitem{bryant2019bea}
C.~Bryant, M.~Felice, {\O}.~E. Andersen, and T.~Briscoe, ``{The BEA-2019 Shared Task on Grammatical Error Correction},'' in \emph{Proceedings of the fourteenth workshop on innovative use of NLP for building educational applications}, 2019, pp. 52--75.

\bibitem{multigec2024}
E.~Volodina, A.~Masciolini, A.~Caines \emph{et~al.}, ``{Shared task on Multilingual Grammatical Error Correction 2025},'' \url{https://www.aclweb.org/portal/content/shared-task-multilingual-grammatical-error-correction-2025}, 2025.

\bibitem{lu2020spoken}
Y.~Lu, M.~J.~F. Gales, and Y.~Wang, ``{Spoken Language ‘Grammatical Error Correction’},'' in \emph{Proc. INTERSPEECH 2020}, 2020, pp. 3840--3844.

\bibitem{qian2025scaling}
M.~Qian, R.~Ma, S.~Bann{\`o}, K.~M. Knill, and M.~J. Gales, ``{Scaling and Prompting for Improved End-to-End Spoken Grammatical Error Correction},'' in \emph{Proc. INTERSPEECH 2025}, 2025.

\bibitem{qian2025sgec-journal}
M.~Qian, R.~Ma, S.~Bann{\`o}, M.~J. Gales, and K.~M. Knill, ``{End-to-End Spoken Grammatical Error Correction},'' \emph{arXiv preprint arXiv:2506.18532}, 2025.

\bibitem{grundkiewicz-junczys-dowmunt-2019-minimally}
\BIBentryALTinterwordspacing
R.~Grundkiewicz and M.~Junczys-Dowmunt, ``{Minimally-Augmented Grammatical Error Correction},'' in \emph{Proceedings of the 5th Workshop on Noisy User-generated Text (W-NUT 2019)}, W.~Xu, A.~Ritter, T.~Baldwin, and A.~Rahimi, Eds.\hskip 1em plus 0.5em minus 0.4em\relax Hong Kong, China: Association for Computational Linguistics, Nov. 2019, pp. 357--363. [Online]. Available: \url{https://aclanthology.org/D19-5546/}
\BIBentrySTDinterwordspacing

\bibitem{10.1016/j.jksuci.2023.101572}
\BIBentryALTinterwordspacing
A.~Solyman, M.~Zappatore, W.~Zhenyu, Z.~Mahmoud, A.~Alfatemi, A.~O. Ibrahim, and L.~A. Gabralla, ``{Optimizing the impact of data augmentation for low-resource grammatical error correction},'' \emph{J. King Saud Univ. Comput. Inf. Sci.}, vol.~35, no.~6, Jun. 2023. [Online]. Available: \url{https://doi.org/10.1016/j.jksuci.2023.101572}
\BIBentrySTDinterwordspacing

\bibitem{koyama2021comparison}
\BIBentryALTinterwordspacing
A.~Koyama, K.~Hotate, M.~Kaneko, and M.~Komachi, ``{Comparison of Grammatical Error Correction Using Back-Translation Models},'' in \emph{Proceedings of the 2021 Conference of the North American Chapter of the Association for Computational Linguistics: Student Research Workshop}.\hskip 1em plus 0.5em minus 0.4em\relax Online: Association for Computational Linguistics, Jun. 2021, pp. 126--135. [Online]. Available: \url{https://aclanthology.org/2021.naacl-srw.16/}
\BIBentrySTDinterwordspacing

\bibitem{luhtaru2024toerr}
A.~Luhtaru, T.~Purason, M.~Vainikko, M.~Del, and M.~Fishel, ``{To Err Is Human, but Llamas Can Learn It Too},'' in \emph{Findings of the Association for Computational Linguistics: EMNLP 2024}.\hskip 1em plus 0.5em minus 0.4em\relax Miami, Florida, USA: Association for Computational Linguistics, Nov. 2024, pp. 12\,466--12\,481.

\bibitem{passali2025artificial}
T.~Passali, T.~Mavropoulos, G.~Tsoumakas, G.~Meditskos, and S.~Vrochidis, ``{Artificial disfluency detection, uh no, disfluency generation for the masses},'' \emph{Computer Speech and Language}, vol.~89, p. 101711, 2025.

\bibitem{dubov2023thesis}
S.~Dubov, ``{Automatic Assessment of English as a Second Language},'' Master's thesis, University of Cambridge, Cambridge, UK, 2021.

\bibitem{huang2023textgenerationspeechsynthesis}
\BIBentryALTinterwordspacing
Z.~Huang, G.~Keren, Z.~Jiang, S.~Jain, D.~Goss-Grubbs, N.~Cheng, F.~Abtahi, D.~Le, D.~Zhang, A.~D'Avirro, E.~Campbell-Taylor, J.~Salas, I.-E. Veliche, and X.~Chen, ``{Text Generation with Speech Synthesis for ASR Data Augmentation},'' 2023. [Online]. Available: \url{https://arxiv.org/abs/2305.16333}
\BIBentrySTDinterwordspacing

\bibitem{Fazel2021}
\BIBentryALTinterwordspacing
A.~Fazel, W.~Yang, Y.~Liu, R.~Barra-Chicote, Y.~Meng, R.~Maas, and J.~Droppo, ``{SynthASR: Unlocking synthetic data for speech recognition},'' 2021. [Online]. Available: \url{https://www.amazon.science/publications/synthasr-unlocking-synthetic-data-for-speech-recognition}
\BIBentrySTDinterwordspacing

\bibitem{knill2024speakimprovecorpus}
\BIBentryALTinterwordspacing
K.~Knill, D.~Nicholls, M.~J. Gales, M.~Qian, and P.~Stroinski, ``{Speak \& Improve Corpus 2025: an L2 English Speech Corpus for Language Assessment and Feedback},'' 2024. [Online]. Available: \url{https://doi.org/10.17863/CAM.114333}
\BIBentrySTDinterwordspacing

\bibitem{lewis2020bart}
M.~Lewis, Y.~Liu, N.~Goyal, M.~Ghazvininejad, A.~Mohamed, O.~Levy, V.~Stoyanov, and L.~Zettlemoyer, ``{BART: Denoising Sequence-to-Sequence Pre-training for Natural Language Generation, Translation, and Comprehension},'' in \emph{Proceedings of the 58th Annual Meeting of the Association for Computational Linguistics}, 2020, pp. 7871--7880.

\bibitem{qian2024speakimprovechallenge}
\BIBentryALTinterwordspacing
M.~Qian, K.~Knill, S.~Banno, S.~Tang, P.~Karanasou, M.~J. Gales, and D.~Nicholls, ``{Speak \& Improve Challenge 2025: Tasks and Baseline Systems},'' 2024. [Online]. Available: \url{https://arxiv.org/abs/2412.11985}
\BIBentrySTDinterwordspacing

\bibitem{bryant-etal-2017-automatic}
C.~Bryant, M.~Felice, and T.~Briscoe, ``{Automatic Annotation and Evaluation of Error Types for Grammatical Error Correction},'' in \emph{Proceedings of the 55th Annual Meeting of the Association for Computational Linguistics (Volume 1: Long Papers)}, R.~Barzilay and M.-Y. Kan, Eds.\hskip 1em plus 0.5em minus 0.4em\relax Vancouver, Canada: Association for Computational Linguistics, Jul. 2017, pp. 793--805.

\bibitem{chen2024f5ttsfairytalerfakesfluent}
\BIBentryALTinterwordspacing
Y.~Chen, Z.~Niu, Z.~Ma, K.~Deng, C.~Wang, J.~Zhao, K.~Yu, and X.~Chen, ``{F5-TTS: A Fairytaler that Fakes Fluent and Faithful Speech with Flow Matching},'' 2024. [Online]. Available: \url{https://arxiv.org/abs/2410.06885}
\BIBentrySTDinterwordspacing

\bibitem{peebles2023scalable-diffusion}
W.~Peebles and S.~Xie, ``Scalable diffusion models with transformers,'' in \emph{Proceedings of the IEEE/CVF international conference on computer vision}, 2023, pp. 4195--4205.

\bibitem{peng2024voicecraft}
P.~Peng, P.-Y. Huang, S.-W. Li, A.~Mohamed, and D.~Harwath, ``{VoiceCraft: Zero-Shot Speech Editing and Text-to-Speech in the Wild},'' in \emph{Proceedings of the 62nd Annual Meeting of the Association for Computational Linguistics (Volume 1: Long Papers)}, 2024, pp. 12\,442--12\,462.

\bibitem{zhang2023speakforeignlanguagesvoice}
\BIBentryALTinterwordspacing
Z.~Zhang, L.~Zhou, C.~Wang, S.~Chen, Y.~Wu, S.~Liu, Z.~Chen, Y.~Liu, H.~Wang, J.~Li, L.~He, S.~Zhao, and F.~Wei, ``{Speak Foreign Languages with Your Own Voice: Cross-Lingual Neural Codec Language Modeling},'' 2023. [Online]. Available: \url{https://arxiv.org/abs/2303.03926}
\BIBentrySTDinterwordspacing

\bibitem{casanova2022yourtts}
\BIBentryALTinterwordspacing
E.~Casanova, J.~Weber, C.~D. Shulby, A.~C. Junior, E.~G{\"o}lge, and M.~A. Ponti, ``{Y}our{TTS}: Towards zero-shot multi-speaker {TTS} and zero-shot voice conversion for everyone,'' in \emph{Proceedings of the 39th International Conference on Machine Learning}, ser. Proceedings of Machine Learning Research, vol. 162.\hskip 1em plus 0.5em minus 0.4em\relax PMLR, 17--23 Jul 2022, pp. 2709--2720. [Online]. Available: \url{https://proceedings.mlr.press/v162/casanova22a.html}
\BIBentrySTDinterwordspacing

\bibitem{casanova24xtts}
E.~Casanova, K.~Davis, E.~Gölge, G.~Göknar, I.~Gulea, L.~Hart, A.~Aljafari, J.~Meyer, R.~Morais, S.~Olayemi, and J.~Weber, ``{XTTS: a Massively Multilingual Zero-Shot Text-to-Speech Model},'' in \emph{Interspeech 2024}, 2024, pp. 4978--4982.

\bibitem{canopy2025orpheus-tts}
T.~C.~L. Team, ``{Towards Human-Sounding TTS},'' \url{https://canopylabs.ai/model-releases}, 2025, accessed: March 29, 2025.

\bibitem{bredin2023pyannote}
H.~Bredin, ``pyannote.audio 2.1 speaker diarization pipeline: principle, benchmark, and recipe,'' in \emph{Interspeech 2023}, 2023, pp. 1983--1987.

\bibitem{banno2024towards}
S.~Bann{\`o}, R.~Ma, M.~Qian, K.~M. Knill, and M.~J. Gales, ``{Towards End-to-End Spoken Grammatical Error Correction},'' in \emph{IEEE International Conference on Acoustics, Speech and Signal Processing (ICASSP)}.\hskip 1em plus 0.5em minus 0.4em\relax IEEE, 2024, pp. 10\,791--10\,795.

\bibitem{ludlow2020official}
K.~Ludlow, \emph{{Official Quick Guide to Linguaskill}}.\hskip 1em plus 0.5em minus 0.4em\relax Cambridge: Cambridge University Press \& Assessment, 2020.

\bibitem{nicholls23_interspeech}
D.~Nicholls, K.~M. Knill, M.~J.~F. Gales, A.~Ragni, and P.~Ricketts, ``{Speak \& Improve: {L2} English Speaking Practice Tool},'' in \emph{Proc. INTERSPEECH 2023}, 2023, pp. 3669--3670.

\bibitem{cefr2001}
\BIBentryALTinterwordspacing
{Council of Europe}, \emph{Common European Framework of Reference for Languages: Learning, Teaching, Assessment}.\hskip 1em plus 0.5em minus 0.4em\relax Cambridge: Cambridge University Press, 2001. [Online]. Available: \url{https://rm.coe.int/1680459f97}
\BIBentrySTDinterwordspacing

\end{thebibliography}

\end{document}